\documentclass[a4paper]{styles/svproc}
\usepackage{url}
\usepackage{footmisc}
\usepackage{type1cm}
\usepackage{multicol}

\usepackage{placeins} %for floatbarriers
\usepackage[unicode]{hyperref} % For referrences
\usepackage{graphicx} %Input Image
\usepackage{wrapfig} %for wrapfigure
\usepackage{tikz}
\usetikzlibrary{shapes, arrows, positioning,fit, shapes.multipart}

\usepackage{amsmath} % for math
\usepackage{amssymb} % more math symbols
\usepackage{mathtools}
\usepackage{alphabeta} % for greek characters in maths
\usepackage{empheq} %emphasize eq -> used in boxed equations custom command
\usepackage{xcolor}

\usepackage{algorithm}
\usepackage{algpseudocode}

\usepackage[linewidth=1pt]{mdframed} %begin with \begin{mdframe}
\usepackage{tcolorbox}               %begin with \begin{tcolorbox}
\usepackage{subcaption}
\usepackage{array}

\usepackage{color,soul}
\definecolor{dkgreen}{rgb}{0,0.6,0}
\definecolor{gray}{rgb}{0.5,0.5,0.5}
\definecolor{mauve}{rgb}{0.58,0,0.82}
\definecolor{mygreen}{RGB}{28,172,0} % color values Red, Green, Blue
\definecolor{mylilas}{RGB}{170,55,241}

\newcommand{\vect}[1]{\mathbf{#1}} % for vectors
\newcommand{\matr}[1]{\mathbf{#1}}

\newcommand{\parder}[2]{\frac{\partial #1}{\partial #2 } } % for partial derivative
\newcommand{\norm}[1]{\left\lVert#1\right\rVert}

\begin{document}
\mainmatter              % start of a contribution
\title{Modeling and In-flight Torso Attitude
Stabilization of a Jumping Quadruped}
\titlerunning{Attitude control of jumping quadruped}  % abbreviated title (for running head)
\author{Michail Papadakis\inst{1} \and Jørgen Anker Olsen\inst{2} \and
Ioannis Poulakakis\inst{1} \and
Kostas Alexis\inst{2}}
\authorrunning{Michail Papadakis et al.} % abbreviated author list (for running head)
\tocauthor{Michail Papadakis, Jørgen Anker Olsen, Ioannis Poulakakis, and Kostas Alexis}
\institute{National Technical University of Athens, 15772 Athens, Greece,\\
\email{michael.d.papadakis@gmail.com}
% ,\\ WWW home page:
% \texttt{http://users/\homedir iekeland/web/welcome.html}
\and
Norwegian University of Science and Technology, \\
Høgskoleringen 1, 7034 Trondheim}

\maketitle              % typeset the title of the contribution

\begin{abstract}

This paper addresses the modeling and attitude control of jumping quadrupeds in low-gravity environments. First, a convex decomposition procedure is presented to generate high-accuracy and low-cost collision geometries for quadrupeds performing agile maneuvers. A hierarchical control architecture is then investigated, separating torso orientation tracking from the generation of suitable, collision-free, corresponding leg motions. Nonlinear Model Predictive Controllers (NMPCs) are utilized in both layers of the controller. 
To compute the necessary leg motions, a torque allocation strategy is employed that leverages the symmetries of the system to avoid self-collisions and simplify the respective NMPC.
%The second layer further employs a torque allocation strategy that leverages the symmetries of the system to avoid self-collisions and allow for a simplified formulation of the optimization problem. 
To plan periodic trajectories online, a Finite State Machine (FSM)-based weight switching strategy is also used. The proposed controller is first evaluated in simulation, where $90^{\circ}$ rotations in roll, pitch, and yaw are stabilized in $6.3$, $2.4$, and $5.5$ seconds, respectively. The performance of the controller is further experimentally demonstrated by stabilizing constant and changing orientation references. Overall, this work provides a framework for the development of advanced model-based attitude controllers for jumping legged systems.

% We would like to encourage you to list your keywords within
% the abstract section using the \keywords{...} command.
\end{abstract}
\section{Introduction}
In recent years, robotic explorers with increasingly advanced capabilities have been proposed for space exploration across various environments, including asteroid surfaces~\cite{SpaceHopper}, the icy terrain of Enceladus~\cite{vaquero2024eels}, and Martian lava tubes~\cite{olsen2023martian}. The latter are of particular interest due to their scientific value (e.g. access to pristine Martian bedrock) and practical advantages, including their ability to provide shielding from radiation and regolith dust \cite{LavaTubesReport}. Exploring these caves presents significant challenges due to their anticipated vast size and complex geometry. To address these difficulties, jumping systems have been proposed as a viable solution. One particularly promising approach is a jumping quadruped capable of not only walking but also jumping, and stabilizing its attitude mid-flight~\cite{olsen2023martian}. Key to this approach is that the low gravity of Mars ($g_{\textrm{Mars}}=3.71~\textrm{m/s}^2$) enables powerful jumps, making it possible to overcome complex obstacles.

\begin{figure}[ht]
    \centering
    \vspace{-1ex}
    \includegraphics[width=1\linewidth]{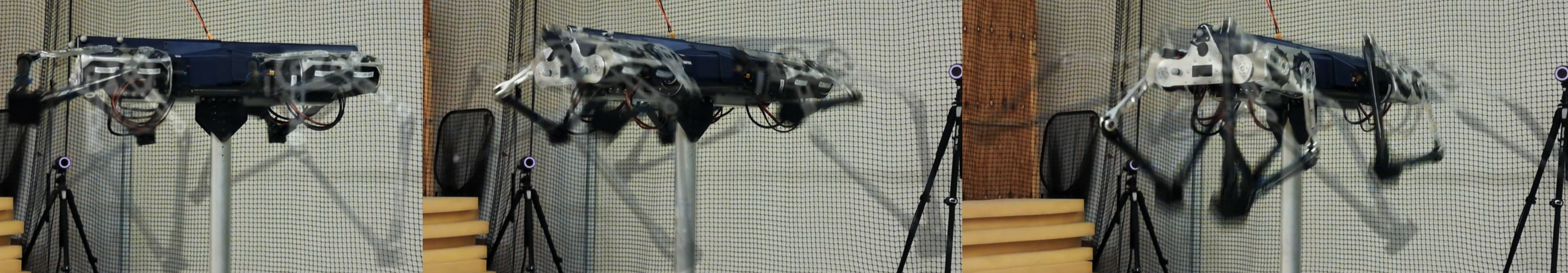}
    \vspace{-4ex}
    \caption{Instances of Olympus changing its yaw orientation by moving its legs  }
    \label{fig:introduction}
\end{figure}

To this end, this work focuses on the modeling and development of a control\-ler for the in-flight attitude stabilization of such a jumping quadruped, namely ``Olympus'', originally presented in~\cite{olsen2023martian}. The proposed controller has a hie\-rarchical structure. Initially, it computes the torso's stabilizing trajectory using a Non\-linear Model Predictive Controller (NMPC). Then it determines the necessary limb motions to reorient the body. More precisely, a torque-tracking NMPC is used to plan the motion of a single leg. Optimizing the trajectory of only one leg is possible, because specific mappings between the motion of that leg and the rest are provided by an allocation strategy. This approach also ensures that no collisions occur between different legs. Furthermore, in order to plan periodic trajectories online, the MPC is wrapped by a Finite State Machine (FSM). To evaluate the proposed strategy for in-flight stabilization, a set of simulation and experimental studies are conducted. Fig.~\ref{fig:introduction} presents an instance from such experiments. The code for this work will be released in:

\noindent \url{https://michalispapadakis.github.io/mpc_olympus}.

The remainder of this paper is organized as follows: 
Section \ref{sec::RelatedWork} outlines related work. Section \ref{sec::Modelling} provides important details regarding the modeling procedure, followed by a detailed description of the control strategy in Sect.~\ref{sec::Control}. Section \ref{sec::Results} presents the evaluation in simulation and real-world experiments, while Sect.~\ref{sec::Conclusion} summarizes and concludes the presented work.

\section{Related Work} \label{sec::RelatedWork}

Several legged systems designed for planetary exploration currently exist, each capable of controlling their attitude mid-flight. SpaceBok is a quadrupedal robot that utilizes a four-bar parallel motion linkage with parallel springs and  can jump up to $1.05$~m ~\cite{Spacebok1}. Initially, it implemented a reaction wheel to control its pitch in the flight phase through PID control. In later iterations, deep reinforcement learning (DRL) was used to stabilize the robot's attitude solely by moving its legs. In simulations, with an additional hip abduction/adduction degree of freedom, it can perform any rotation in less than $3$~s and re-orient itself for a safe landing~\cite{Spacebok2}. 
SpaceHopper is a three-legged, CubeSat-sized,  lightweight robot tailored to low-gravity locomotion, such as on asteroids and moons. It also utilizes DRL to control its orientation. In simulations, it performs an approximately $90^{\circ}$ composite reorientation maneuver in $1$~s  in zero gravity. In experiments on Earth, where it  was attached to a gimbal test stand, it executed similar rotational maneuvers in $5$~s~\cite{SpaceHopper}.

Olympus is a quadrupedal robot that employs a five-bar design for its legs with springs for improved jumping. A two-leg prototype could reach jump heights of $1.52$~m in Earth's gravity and $3.63$~m in Mars' gravity \cite{OlsenDesign}. The first iteration of Olympus employed a hierarchical controller to regulate its attitude in simulation. An MPC calculates the necessary body torques to stabilize the robot, and then a torque allocation algorithm determines the motions of the legs that produce these torques. Simulation results demonstrate the robot's ability to stabilize its attitude in $3$~s on average  \cite{westre2024model}. However, this approach does not consider the full dynamics, treating the legs merely as contracting pendulums, and does not handle self-collisions. 

Although not exclusively designed for planetary exploration, the Mini Cheetah quadruped managed to perform a $90^{\circ}$ pitch maneuver in $0.5$~s. Supervised learning was used to map the robot's initial condition to an optimal trajectory, determined offline through trajectory optimization \cite{cheetah_falling}.

More broadly, the attitude control of a free-floating, rigid body has been extensively studied for space applications. Typically, specialized actuators (e.g. magnetorquers) are used \cite{Nasa_ADCS}. However, theoretical approaches that rely solely on onboard manipulators have also been developed. In \cite{virtual_manipulator}, attitude control is achieved using repetitive circular motions, though these produce only minor effects. Alternatively, pre-planned large manipulator motions can control a spacecraft's orientation, but such methods typically do not consider collision constraints \cite{manipulator_spacecraft}. This issue is similar to the falling cat problem, where internal torques cause reorientation maneuvers. It is noted that in the general form of the falling cat problem, the system has extra degrees of freedom in its waist \cite{falling_cat}.

The main contributions of this work include a detailed and computationally efficient collision modeling process and the development of a torso attitude controller. The modeling procedure allows for efficient collision handling and mathematical formulation of workspace constraints. Furthermore, the model-based nature of the controller makes it more amenable to formal verification, which is particularly important for space systems. Finally, the viability of the proposed controller is evaluated both in simulations and experiments.  \label{sec::Intro}

\section{Modeling} \label{sec::Modelling}
In this section, the dynamic models that are used in the hierarchical controllers are initially presented. Next, the collision modeling procedure is detailed.
\subsection{Dynamics} \label{Modelling::sec::Dynamics}
The first modeling abstraction involves solely the robot's torso which is treated as a rotating rigid body. Only its orientation is of interest, which is parameterized by quaternions $\vect{q} = [q_{\mathrm{w}}, \vect{u}_{\mathrm{q}}^T]^T \triangleq   [q_{\mathrm{w}},q_{\mathrm{x}},q_{\mathrm{y}},q_{\mathrm{z}}]^T $, using the Hamilton convention~\cite{quaternion_kf}, to avoid singularities. The full state of the model is $\vect{x}_{\mathrm{B}} = [\vect{q}^T, \boldsymbol{\omega}^T ]^T$, where $\boldsymbol{\omega}$ is the angular velocity. The dynamics take the form \cite{quaternion_kf,Goldstein2001-iq}:

\begin{equation} %checked michalis
    \dot{\vect{x}}_{\mathrm{B}} \triangleq \vect{f}_{\mathrm{B}} = \begin{bmatrix} \frac{1}{2}\boldsymbol{\Omega}\vect{q} \\  \matr{I}_\mathrm{B}^{-1} \left ( \boldsymbol{\tau}_\mathrm{B} -  \boldsymbol{\omega} \times \matr{I}_\mathrm{B} \boldsymbol{\omega} \right ) \end{bmatrix}\;, \ \text{where $ \boldsymbol{\Omega} = 
    \begin{bmatrix} 
      0                 &  -\omega_\mathrm{x}   &   -\omega_\mathrm{y}  &  -\omega_\mathrm{z} \\  
    \omega_\mathrm{x}   &     0                 &    \omega_\mathrm{z}  &  -\omega_\mathrm{y} \\
    \omega_\mathrm{y}   &  -\omega_\mathrm{z}   &           0           &   \omega_\mathrm{x} \\
    \omega_\mathrm{z}   &   \omega_\mathrm{y}   &   -\omega_\mathrm{x}  &     0   
    \end{bmatrix}\;,$}
    \label{MPC::eq::StateSpaceRotationDynamics}
\end{equation}
where $\matr{I}_{\mathrm{B}}$ is the moment of inertia and $\boldsymbol{\tau}_{\mathrm{B}}$ are the external torques, both expressed in a fixed, robot-centered frame   $\{B\}$.

\begin{wrapfigure}[15]{r}{0.49\textwidth}
    \centering
    \vspace{-6ex}
    \includegraphics[width= 0.4\textwidth]{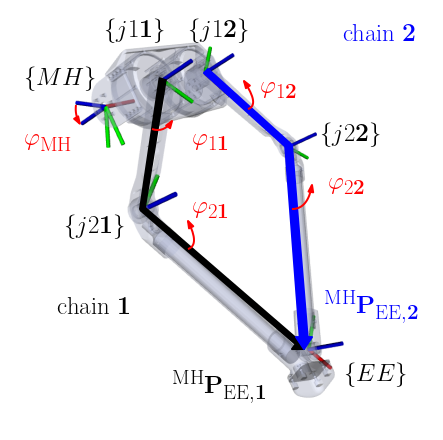}
    \vspace{-1.5ex}
    \caption{Olympus's leg with the corresponding frames}
    \label{Modelling::fig::LegFig}
\end{wrapfigure}

Next, the model of a single leg is considered. Olympus's leg is depicted in Fig. \ref{Modelling::fig::LegFig} along with the definitions of the corresponding joint angles $\varphi_i$. The kinematic analysis of five-bar mechanisms is well studied in literature~\cite{5-bar_kinematics} and is thus omitted here. 

The leg contains a closed kinematic chain that is expressed by the following constraint:

\begin{equation} % OK
    \vect{h} =\ ^{\mathrm{MH}}\vect{P}_{\mathrm{EE},1} - \ ^{\mathrm{MH}}\vect{P}_{\mathrm{EE},2} = \vect{0} \ ,
    \label{Modelling::eq::ClosureConstraint}
\end{equation}

\noindent where $^{\mathrm{MH}}\vect{P}_{\mathrm{EE},i}$ is the position of the end effector ($EE$) expressed with the coordinates of the $i$-th kinematic chain, in the motor-housing frame $\{MH\}$. The dynamics of a single leg are given by~\cite{underactuated}:

\begin{subequations}\label{Modelling::eq::LegDynamics}
\begin{equation} %checked michalis
    \matr{M}(\boldsymbol{\varphi}) \ddot{\boldsymbol{\varphi}} + \matr{C}(\boldsymbol{\varphi},\dot{\boldsymbol{\varphi}}) \dot{\boldsymbol{\varphi}} +\matr{D} \dot{\boldsymbol{\varphi}} + \vect{G}(\boldsymbol{\varphi}) = \boldsymbol{\tau} + \parder{\vect{h}} {\boldsymbol{\varphi}}^{T} \boldsymbol{\lambda}\; ,
    \label{Modelling::eq::LagrangeEOM_constr}
\end{equation}
\begin{equation} %checked michalis
    \boldsymbol{\lambda} = - (\matr{H}\matr{M}^{-1}\matr{H}^{T})^{-1} \left (\matr{H}\matr{M}^{-1} ( \boldsymbol{\tau}  -\matr{C}\dot{\boldsymbol{\varphi}} - \matr{D}\dot{\boldsymbol{\varphi}} -\matr{G} ) +\matr{\dot{H}} \boldsymbol{\varphi} \right)\; .
    \label{Modelling::eq::LagrangeEOM_lambda}
\end{equation}
\end{subequations}
where $\boldsymbol{\varphi} = [\varphi_{\mathrm{MH}},\ \varphi_{\mathrm{11}},\ \varphi_{\mathrm{21}},\ \varphi_{\mathrm{12}},\ \varphi_{\mathrm{22}}]^T$ is the configuration of the leg,  $\matr{M}$ is the mass matrix, $\matr{C}\dot{\boldsymbol{\varphi}}$ accounts for Coriolis and centrifugal forces, $\matr{D}\dot{\boldsymbol{\varphi}}$ accounts for the viscous damping forces, $\vect{G}$ contains the gravity forces, $\boldsymbol{\tau}$ contains the motor torques, $\boldsymbol{\lambda}$ is the closure constraint force vector and $\matr{H} \triangleq \partial\vect{h}/ \partial \boldsymbol{\varphi}$ is the jacobian of the constraints. Expression (\ref{Modelling::eq::LagrangeEOM_lambda}) requires that $ (\matr{H}\matr{M}^{-1}\matr{H}^{T})$ is invertible. This holds if $\vect{H}$ is full rank \cite{FeatherstoneDynamics}, which was checked with a $200\times200$ grid search over the involved independent angles $\varphi_{\mathrm{11}}$ and $\varphi_{\mathrm{12}}$, sampled uniformly within their respective ranges. 

The frames of reference for both the torso and leg dynamics translate together with the quadruped. During a jump, the whole robot experiences the same gravitational acceleration, so, in these frames of reference, the relative acceleration of the links due to gravity is zero. Therefore, $\vect{G} = \vect{0}$ in (\ref{Modelling::eq::LagrangeEOM_constr}) and (\ref{Modelling::eq::LagrangeEOM_lambda}). Also, the rotational behavior of the robot during jumping can be studied at free floating conditions without gravity in simulation.

\subsection{Collision Modeling and Workspace Constraints}

\begin{figure}
    \centering
    \vspace{-3ex}
    \includegraphics[width=0.95\linewidth]{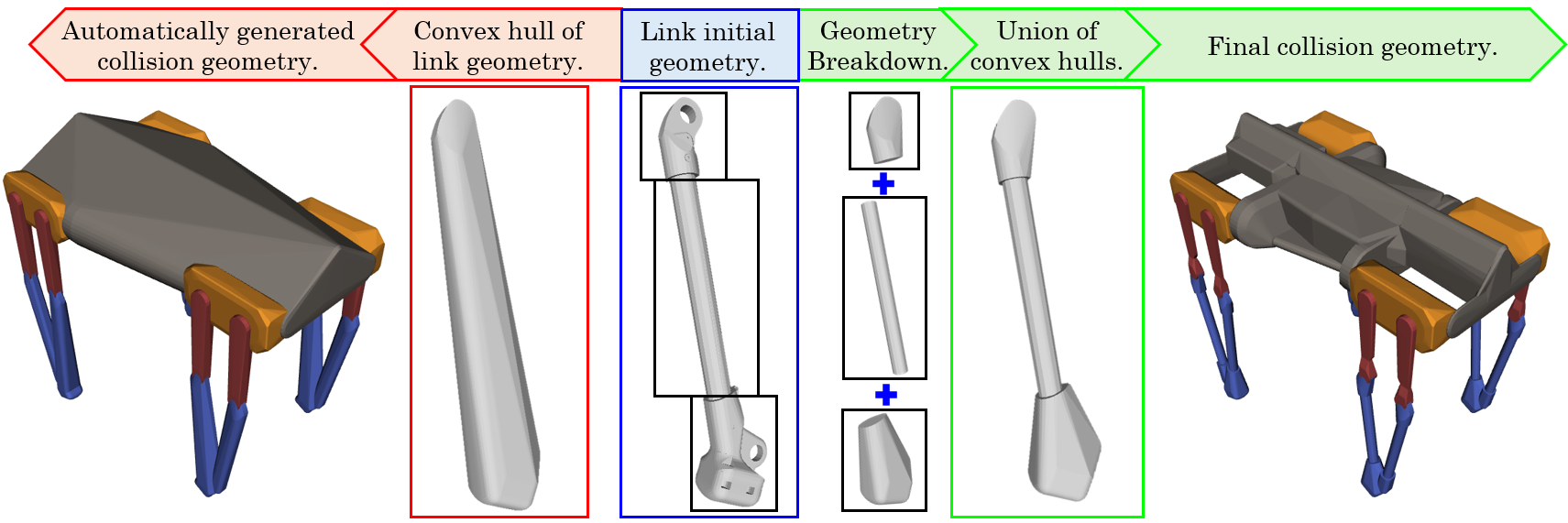}
    \vspace{-1ex}
    \caption{Procedure for generating the collision geometry}
    \vspace{-3ex}
    \label{Modelling::fig::collision_geometry}
\end{figure}

Reorientation maneuvers often bring the limbs close to the torso, requiring detailed contact modeling to capture the full range of motions in simulation. Both Matlab and Drake~\cite{drake}, which are employed in this work for prototyping and evaluation studies respectively, use the convex hull of the link geometries for efficient contact 
handling. The result is very coarse and restrictive for our purpose, as illustrated in Fig. \ref{Modelling::fig::collision_geometry}. To address this, the link geometries are decomposed into a union of smaller convex shapes, offering a more accurate yet simplified representation of their collision geometry. The convexification process along with the resulting geometry are shown in Fig.  \ref{Modelling::fig::collision_geometry}.

A grid search over actuated angles $\varphi_{\mathrm{MH}},\ \varphi_{\mathrm{11}},\ \varphi_{\mathrm{12}}$ was used to identify colliding configurations, from which polytopic (\ref{Modelling::eq::polytopic_state_only}) and nonlinear (\ref{Modelling::eq::general_state_constraint}) constraints were manually extracted for use in the controller presented in Sec. \ref{sec::Control}. 

\begin{equation}
    \matr{C} \boldsymbol{\varphi} < \vect{c}\; , \label{Modelling::eq::polytopic_state_only}
\end{equation}
\begin{equation}
    \vect{G} (\boldsymbol{\varphi}) =  \begin{bmatrix}
    P_5(\varphi_{\mathrm{MH}}) + \varphi_{\mathrm{12}}  \\ 
    \sum_{0}^{1}σ_{log,i}(\varphi_{\mathrm{MH}}) y_i(\varphi_{\mathrm{MH}})  +\varphi_{\mathrm{11}}
\end{bmatrix} < \vect{g} . 
    \label{Modelling::eq::general_state_constraint}
\end{equation}
where $y_i$ is a linear function, $σ_{log,i}$ is a logistic function and $P_5$ is a fifth degree polynomial. These constraints are non convex but can be handled by acados \cite{acados},  the employed MPC framework. 

\section{Model Predictive Orientation Control} \label{sec::Control}
The goal of the controller is to stabilize the attitude of the quadruped’s torso and reach the desired orientation by moving the legs of the robot. It features a hierarchical structure with two distinct components: the Body Planner and the Leg Planner. The Body Planner calculates the torso's trajectory to track a reference orientation, while the Leg Planner determines the leg motions necessary to execute the reorientation maneuver. Both components primarily rely on NMPCs, but the Leg Planner also incorporates a torque allocation module and a resetting strategy.

In Sects. \ref{MPC::sec::BodyPlanner}--\ref{MPC::sec::Allocation} key elements of the controller are analyzed, while in Sect. \ref{MPC::sec::CompleteArch} the complete architecture is outlined. Figure \ref{MPC::fig::ControllerArchitecture} in Sect. \ref{MPC::sec::CompleteArch} provides a schematic overview of the controller. 

\subsection{Body Planner} \label{MPC::sec::BodyPlanner}
To calculate the torso reorientation maneuvers, the body planner treats the robot as a single rotating rigid body, considering the legs as torque-producing actuators. Note that the quaternion error $\vect{q}_\mathrm{e}$ from the current ($\vect{q}$) to the desired ($\vect{q}_\mathrm{D}$) orientation is given by: 

\begin{equation} %ok michalis
    \vect{q}_\mathrm{e}  = \vect{q}_\mathrm{D} \otimes \vect{q}^{-1}\; ,
    \label{MPC::eq::QuaternionError}
\end{equation}
where $\otimes$ denotes the Hamilton product and $\vect{q}^{-1}$ is the inverse quaternion ~\cite{quaternion_kf}. The angle axis representation of the orientation error is given by:
\begin{equation}
    \vect{f}_{\mathrm{q}} = \theta \vect{u}_{\mathrm{e}}\;,
    \label{MPC::eq::QuaternionMetric}
\end{equation}
where $\vect{u}_{\mathrm{e}} =  \vect{u}_{\mathrm{q,e}} / || \vect{u}_{\mathrm{q,e}} ||$ and $\theta$ is given by:
\begin{equation}
    \theta = \arccos \left (\  2   (\vect{q}_{\mathrm{D}} \cdot   \vect{q} ) ^2 \ -\ 1 \ \right )\;.
\end{equation}

Building on this analysis, the Body Planner MPC is established\footnote{
Here $\norm{ \vect{v} }_\matr{P}^2 = \vect{v}^T \matr{P} \vect{v}$ where $\vect{v}$ is a $n\times1$ vector and $\matr{P}$ a $n\times n$ weight matrix, which is positive semi-definite in  (\ref{MPC::eq::BodyPlannerMPC}) and (\ref{MPC::eq::torque_tracking}). All vector quantities in (\ref{MPC::eq::BodyPlannerMPC}) and (\ref{MPC::eq::torque_tracking}) are of size $3$, except for the leg state $\vect{x}$ which is $10 \times 1$.
} :
\begin{equation}
    \min_{\vect{x}_{\mathrm{B}},\boldsymbol{\tau}_{\mathrm{B}} } \int_0^{T_{\mathrm{H,B}}} \left[ 
    \norm{ \vect{f}_{\mathrm{q}}             }^2_{\matr{Q}_\mathrm{q}} + 
    \norm{ \boldsymbol{\omega}      }^2_{\matr{Q}_{\omega}} + 
    \norm{ \boldsymbol{\tau}_{\mathrm{B}}      }^2_{\matr{R}}  
    \right ]dt + 
    \norm{ \vect{f}_{\mathrm{q,E}}           }^2_{\matr{Q}_{\mathrm{q,E}} } + 
    \norm{ \boldsymbol{\omega}_\mathrm{E}    }^2_{\matr{Q}_{\omega,\mathrm{E}} }
    \label{MPC::eq::BodyPlannerMPC}
\end{equation}
subject to

\begin{subequations} \label{MPC::eq::PosControlMPC}
\begin{align}
    \dot{\vect{x}}_{\mathrm{B}}      &= \vect{f}_{\mathrm{B}}( \vect{x}_{\mathrm{B}},\boldsymbol{\tau}_{\mathrm{B}} )\; ,            \label{MPC::eq::dynamic_constr} \\
    \vect{f}_{\mathrm{q,E}}        &= \vect{0}\; ,                                                 \label{MPC::eq::BodyPlannerTC} \\
     \vect{x}_{\mathrm{B},0}       &= \vect{x}_{\mathrm{B}}(t)\;,                                  \label{MPC::eq::BodyPlannerX0} \\
    \boldsymbol{\tau}_{\mathrm{B}} & \in [\boldsymbol{\tau}_{\mathrm{B,l}},\ \boldsymbol{\tau}_{\mathrm{B,u}}]\;,   \label{MPC::eq::BodyPlannerInputConstr}
\end{align}
\end{subequations}
where $\vect{x}_{\mathrm{B}}$ is the state and $\boldsymbol{\tau}_{\mathrm{B}}$ is the input of the system, as defined in Sect. \ref{Modelling::sec::Dynamics}, and  $T_{\mathrm{H,B}}$ is the prediction horizon of the MPC. 
The term $\norm{ \vect{f}_{\mathrm{q}}        }^2_{\matr{Q}_{\mathrm{q}}}$ penalizes orientation errors, while $\norm{ \boldsymbol{\omega}      }^2_{\matr{Q}_{\omega}}$ and $\norm{ \boldsymbol{\tau}_{\mathrm{B}} }^2_{\matr{R}}$ penalize large angular velocities and inputs, respectively. Expressions $\norm{ \vect{f}_{\mathrm{q,E}} }^2_{\matr{Q}_{\mathrm{q,E}}}$and 
$\norm{ \boldsymbol{\omega}_{\mathrm{E}} }^2_{\matr{Q}_{\omega,\mathrm{E}}}$ are their corresponding terminal cost analogs.
    
The dynamics (\ref{MPC::eq::dynamic_constr}) are given by (\ref{MPC::eq::StateSpaceRotationDynamics}). Constraint (\ref{MPC::eq::BodyPlannerInputConstr}) is the bound on input\footnote{
In (\ref{MPC::eq::BodyPlannerInputConstr}) and  (\ref{MPC::eq::leg_state_constr}--\ref{MPC::eq::leg_nonlinear_constr}) the subscripts $l,u$ denote the lower and upper bounds respectively.
},    
and (\ref{MPC::eq::BodyPlannerX0}) sets the initial system state $\vect{x}_{\mathrm{B},0}$ to the current state of the torso $\vect{x}_{\mathrm{B}}(t)$. Equality  (\ref{MPC::eq::BodyPlannerTC}) ensures the desired orientation is reached. It is slacked, as the discretization or input bounds could compromise its satisfaction.

\subsection{Leg Planner: MPC Formulation}
 Due to the allocation strategy, analyzed in Sec. \ref{MPC::sec::Allocation}, the MPC of the Leg Planner calculates only how one leg can induce a torque at its base $\boldsymbol{\tau}_{\mathrm{B}}(\boldsymbol{\varphi},\boldsymbol{\tau})$ to track $\boldsymbol{\tau}_{\mathrm{B}}^{\ast}$ from the Body Planner, employing a detailed dynamic model of the leg (\ref{Modelling::eq::LegDynamics}). The formulation of the Leg Planner MPC is the following:

\small
\begin{equation}
    \min_{\vect{x},\boldsymbol{\tau}} \int_0^{T_{\mathrm{H,L}}} \left[ 
    \norm{\boldsymbol{\tau}_{\mathrm{B}}(\boldsymbol{\varphi},\boldsymbol{\tau}) - \boldsymbol{\tau}_{\mathrm{B}}^*}_{\matr{W}_{\mathrm{tr}}}^2 +  
    \norm{ \boldsymbol{\tau}           }_{\matr{W}_{\tau}}^2 +
    \norm{  \vect{x}-\vect{x}_{\mathrm{ref}}    }_{\matr{Q}^2 }
    \right ]dt + 
    \norm{  \vect{x}_{\mathrm{E}}-\vect{x}_{\mathrm{ref}} }_{\matr{Q}_{\mathrm{E}}}^2 
    \label{MPC::eq::torque_tracking}
\end{equation}
\normalsize
subject to

\begin{subequations}
\begin{align}
    \dot{\vect{x}}        &= \vect{f}( \vect{x},\boldsymbol{\tau} )\;,              \label{MPC::eq::leg_dynamics} \\
    \vect{x}_0            & = \vect{x}(t)\;,                                        \label{MPC::eq::leg_state_constr_x0} \\
     \vect{x}             &\in [\vect{x}_{\mathrm{l}},\ \vect{x}_{\mathrm{u}} ]\;,                    \label{MPC::eq::leg_state_constr} \\
     \boldsymbol{\tau}    &\in [\boldsymbol{\tau}_{\mathrm{l}},\ \boldsymbol{\tau}_{\mathrm{u}} ]\;,  \label{MPC::eq::leg_input_constr} \\
     \matr{C}\vect{x}     &\in [\vect{c}_{\mathrm{l}},\ \vect{c}_{\mathrm{u}} ]\;,                    \label{MPC::eq::leg_linear_constr} \\
     \vect{G}(\vect{x})   &\in [\vect{g}_{\mathrm{l}},\ \vect{g}_{\mathrm{u}} ]\;,                    \label{MPC::eq::leg_nonlinear_constr} \\
\end{align}
    \label{MPC::eq::leg_constr}
\end{subequations}
where  $\vect{x} = [\boldsymbol{\varphi}^{T},\dot{\boldsymbol{\varphi}}^{T}]^T$ is the state and $\boldsymbol{\tau}$ is the input of the system, and  $T_{\mathrm{H,L}}$ is the horizon of the MPC. The term $\norm{\boldsymbol{\tau}_{\mathrm{B}}(\boldsymbol{\varphi},\boldsymbol{\tau}) - \boldsymbol{\tau}_{\mathrm{B}}^*}_{\matr{W}_{\mathrm{tr}}}^2$ penalizes torque tracking deviations, while
$\norm{ \boldsymbol{\tau}}_{\matr{W}_{\tau}}^2$ penalizes large actuation inputs and smooths the solution. The expression
$ \norm{ \vect{x}-\vect{x}_{\mathrm{ref}} }_{\matr{Q}^2 }$ is a state tracking term, and
$\norm{  \vect{x}_{\mathrm{E}}-\vect{x}_{\mathrm{ref}} }_{\matr{Q}_{\mathrm{E}}}^2$ is its terminal cost analog. Here,
$\vect{x}_{\mathrm{ref}} = [\boldsymbol{\varphi}_{\mathrm{ref}}^{T},\vect{0}^T]^T$, where $\boldsymbol{\varphi}_{\mathrm{ref}}$ is selected by the resetting strategy presented in Sect. \ref{MPC::sec::Reset}, and  $\matr{Q} = \textrm{diag}(\matr{Q}_{\boldsymbol{\varphi}},\matr{Q}_{\dot{\boldsymbol{\varphi}}})$ to influence $\boldsymbol{\varphi}$ and $\dot{\boldsymbol{\varphi}}$ respectively.

The dynamics (\ref{MPC::eq::leg_dynamics})  is the state-space representation of (\ref{Modelling::eq::LegDynamics}).  
Constraint  (\ref{MPC::eq::leg_state_constr_x0}) sets the initial system state $\vect{x}_{0}$ to the current state of the leg $\vect{x}(t)$, while  (\ref{MPC::eq::leg_state_constr},\ref{MPC::eq::leg_input_constr}) are the bounds on state and input.  
Constraint (\ref{MPC::eq::leg_linear_constr}) denotes the linear workspace constraints (\ref{Modelling::eq::polytopic_state_only}), while constraint (\ref{MPC::eq::leg_nonlinear_constr}) represents both the non-linear workspace constraints (\ref{Modelling::eq::general_state_constraint}) as well as the kinematic closure constraint (\ref{Modelling::eq::ClosureConstraint}). 
All constraints are slacked to ensure the feasibility of the optimization problem,  apart from (\ref{MPC::eq::leg_dynamics}, \ref{MPC::eq::leg_state_constr_x0}, \ref{MPC::eq::leg_input_constr}). Slacking the dynamics would alter the prediction model while slacking the input is unnecessary as it is an independent variable. Moreover, the dynamics are well defined in the workspace, as discussed in Sec. \ref{Modelling::sec::Dynamics}.

\subsection{Leg Planner: Torque Allocation} \label{MPC::sec::Allocation}

By leveraging the system symmetries, one can solve the optimization problem described by (\ref{MPC::eq::torque_tracking}) and (\ref{MPC::eq::leg_constr}) for one leg and then map the optimal joint trajectories of that leg to the other legs. Careful construction of this conversion can also prevent collisions between different legs, which the current MPC formulation does not account for. The mapping is essentially an affine transformation of the optimal trajectories of one leg's actuated angles to another.

\begin{figure}[ht]
    \centering
    \includegraphics[width=1\linewidth]{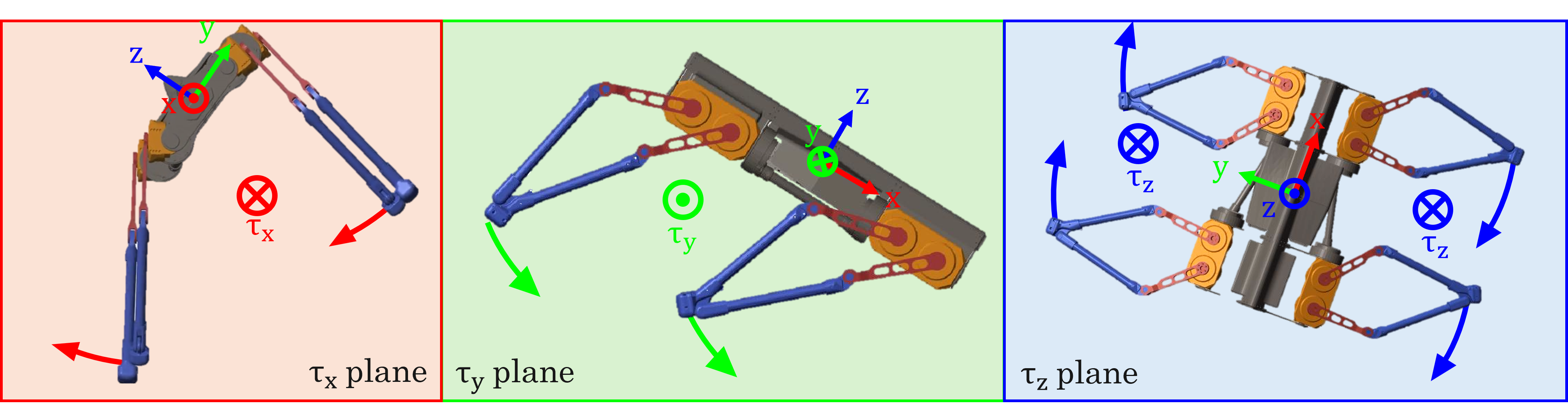}
    \caption{ The curved arrows indicate the direction of the leg movements, while the arrow tips/tails denote the direction of the induced torque. \textbf{Roll degree of freedom}:~If the legs move together in the same direction in the $\tau_{\mathrm{x}}$ plane a net roll torque is induced in the torso (\emph{roll} mapping), whereas if they move in opposite directions in the $\tau_{\mathrm{x}}$ plane, the roll torques almost cancel out (\emph{no roll} mapping).  \textbf{Pitch/Yaw degree of freedom}:~ To produce pitch moments, the legs must move near the $\tau_{\mathrm{y}}$ plane in unison from rear to front or vice versa (\emph{pitch} mapping). To produce yaw moments, the legs of one side move from rear to front  and the ones from the other side from front to rear near the $\tau_{\mathrm{z}}$ plane (\emph{yaw} mapping).  }
    \label{MPC::fig::AllocationModes}
\end{figure}

Firstly, the rear legs mimic the movements of the front ones to avoid collisions between them while using their full workspace. By specifying the opposite-side mapping, the movements of all the legs are coupled. This directly affects the body torques that can be produced by the maneuvers of the legs. Opposite-side mapping has two \emph{degrees of freedom}: whether rolling moments will be produced or not (\emph{roll} and \emph{no roll} mappings), and whether the legs will produce pitch or yaw moments (\emph{pitch} and \emph{yaw} mapping). These are shown in Fig. \ref{MPC::fig::AllocationModes}.

So according to the selected mapping settings, the controller can operate in four different modes:
\begin{itemize}
    \item \textbf{Roll Mode}:  It uses the \textit{roll} and \textit{pitch} mappings.
    \item \textbf{Pitch Mode}: It uses the \textit{no roll} and \textit{pitch} mappings.
    \item \textbf{Yaw Mode}:   It uses the \textit{no roll} and \textit{yaw} mappings.
    \item \textbf{Stabilization Mode}: It uses the \textit{no roll} and inherits the previous \textit{pitch} or \textit{yaw} mapping.
\end{itemize}
The operating mode is selected based on the largest term of $\boldsymbol{\tau}_{\mathrm{B}}^{\ast}$. The controller switches to stabilization mode once the robot's orientation is within a defined angular threshold of the target orientation. 

\subsection{Leg Planner: Resetting Strategy} \label{MPC::sec::Reset}

The short horizon of the leg planner's MPC ($T_{\mathrm{H,L}}=0.1$~s) does not allow it to plan the entire motion on its own. Additionally, when the leg reaches the boundary of the workspace, moving to a \emph{resetting} point---where it can efficiently resume tracking the reference torque---conflicts with the short-term objectives of the MPC. Most probably, this motion will produce  torque in the opposite direction. Thus, a guiding (resetting) strategy must be wrapped around the MPC.

\begin{figure}[ht]
    \centering
    \includegraphics[width= 0.8\linewidth]{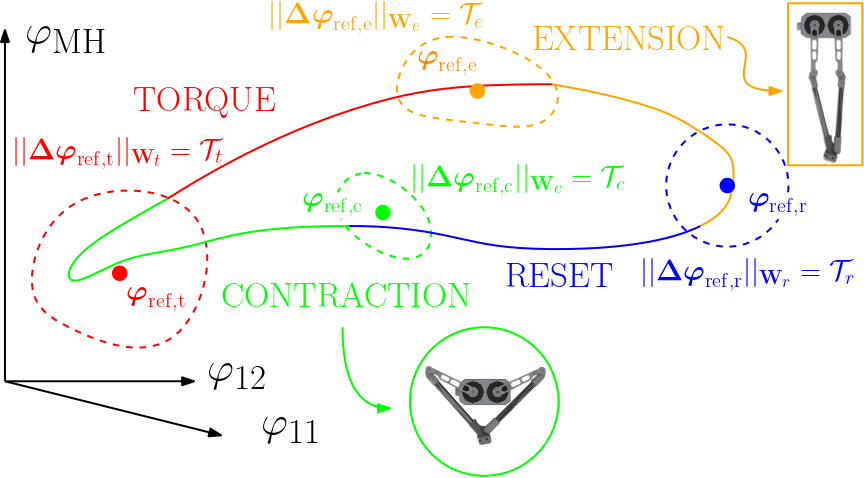}
    \caption{Illustration of the resetting algorithm on an example trajectory. The controller tracks the current phase setpoint ($\boldsymbol{\varphi}_{\mathrm{ref,i}}$). 
    When the current joint position $\boldsymbol{\varphi}(t)$ enters the phase's ‘‘bubble" as per (\ref{MPC::eq::state_trans}), the next phase becomes active}
    \label{MPC::fig::reset_algo}
\end{figure}

A geometrical overview of the resetting algorithm in the actuated joint angle space is shown in Fig. \ref{MPC::fig::reset_algo}. 
Intuitively, there is a phase in the trajectory that produces the desired torque, followed by another one where the leg goes to an advantageous position to continue tracking the torque in the future. These are the \emph{TORQUE} and \emph{RESET} phases, respectively. These phases are infused with two intermediate phases: \emph{EXTENSION} and \emph{CONTRACTION}. In these phases, the whole leg's center of mass moves further or closer to the body to increase or decrease the induced torque in the torso. Furthermore, in the pitch and yaw modes, the legs move towards or away from the pitch and yaw planes indicated in Fig. \ref{MPC::fig::AllocationModes}. 

The resulting resetting algorithm functions as a FSM, where each state/phase is linked with a set of weights for the MPC and a corresponding reference setpoint. The parameters of the FSM are:

\begin{itemize}
    \item A reference setpoint $\boldsymbol{\varphi}_{\mathrm{ref,i}}$ (5 parameters, 3 of them independent).
    \item Weight values $\matr{W}_{\mathrm{MPC}}$ for the Leg Planner (12 parameters\footnote{
    $\matr{W}_{\mathrm{tr}}$ (3 values), $\matr{W}_{\boldsymbol{\tau}}$ (1 value) and $\matr{Q}_{\boldsymbol{\varphi}},\ \matr{Q}_{\boldsymbol{\varphi},E},\ \matr{Q}_{\dot{\boldsymbol{\varphi}}},\ \matr{Q}_{\dot{\boldsymbol{\varphi}},\mathrm{E}}$ (8 values) as different weights are given for $\varphi_{\mathrm{MH}}$ and $\dot{\varphi}_{\mathrm{MH}}$, and the rest of the joint angles.})
    \item A diagonal weight matrix $\matr{W}_{\mathrm{i}}$, and a threshold value $\mathcal{T}_{\mathrm{i}}$ needed for the state transition function (\ref{MPC::eq::state_trans}) (4 parameters). 
\end{itemize}

The state transition function of the FSM is the following:
\begin{equation}
    \norm{ \Delta \boldsymbol{\varphi}_{\mathrm{ref,i}} }_{\matr{W}_{\mathrm{i}}} \triangleq \norm{ \boldsymbol{\varphi}(t) - \boldsymbol{\varphi}_{\mathrm{ref,i}} }_{\matr{W}_{\mathrm{i}}}  \leq \mathcal{T}_{\mathrm{i}} \; .
    \label{MPC::eq::state_trans}
\end{equation}

Each state has 21 parameters, which results in 84 parameters for the whole FSM. Due to the allocation algorithm of Sect.~\ref{MPC::sec::Allocation}, these parameters are different for roll, pitch and yaw. Therefore, the entire controller comprises a total of 252 parameters. These parameters were manually tuned for satisfactory results.

\subsection{Complete architecture} \label{MPC::sec::CompleteArch}

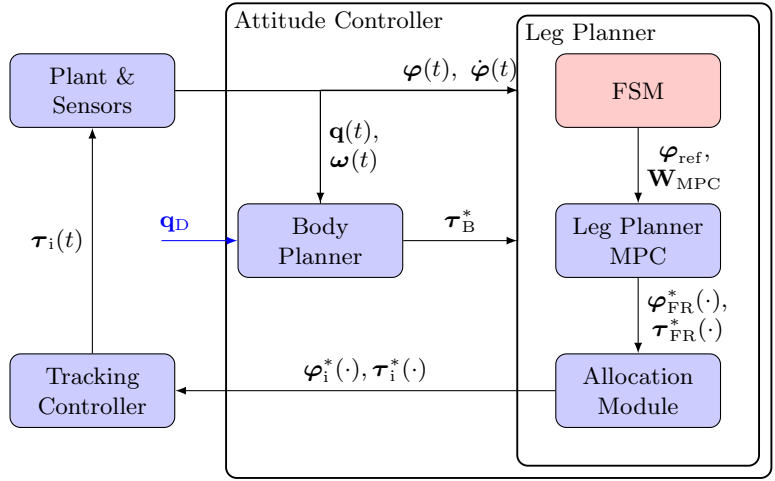
\begin{figure}[ht]
    \centering
    \begin{tikzpicture}[
      block/.style = {rectangle, draw, fill=blue!20, text width=6em, text centered, rounded corners, minimum height=3em},
      input/.style = {coordinate},
      arrow/.style = {draw, -latex}
    ]

      % Nodes
    \node[block] (BP)                    {Body Planner};    
    \node[block, right=2cm of BP] (LP)      {Leg Planner MPC};
    \node[block, above=1cm of LP, fill=red!20] (fsm) {FSM};
    \node[block, below=1cm of LP] (all)  {Allocation Module};
    \node[block, left=5cm of all] (pid)  {Tracking Controller};
    \node[block, left=5cm of fsm] (plant)  {Plant \& Sensors};

    %Supernodes
    \node[draw=black!250, rounded corners,inner sep=0.5cm, fit={(fsm) (LP) (all)}, line width=0.25mm] (LPC) {};
    \node[anchor=north west] at (LPC.north west) {Leg Planner};   

    \node[draw=black!250, rounded corners, inner sep=0.15cm,  fit={ (LPC) (BP) },line width=0.25mm] (HLC) {};
    \node[anchor=north west] at (HLC.north west) {Attitude Controller};

    % Arrows
    \draw[arrow] (BP.east)  --  ++(1.5,0.0) node[midway, above] {$\boldsymbol{\tau}_{\mathrm{B}}^{\ast} $} (LPC);
    \draw[arrow] (fsm) --node[midway, right] {$\begin{matrix} \boldsymbol{\varphi}_{\mathrm{ref}} , \\ \matr{W}_{\mathrm{MPC}}\end{matrix}$}   (LP);
    \draw[arrow] (LP)  --node[midway, right] {$\begin{matrix} \boldsymbol{\varphi}_{\mathrm{FR}}^{\ast}(\cdot) ,\\ \boldsymbol{\tau}_{\mathrm{FR}}^{\ast}(\cdot) \end{matrix}$}   (all);
    \draw[arrow] (all) --node[midway, above] {$\boldsymbol{\varphi}_{\mathrm{i}}^{\ast}(\cdot), \boldsymbol{\tau}_{\mathrm{i}}^{\ast}(\cdot) $}   (pid);
    \draw[arrow] (pid) --node[midway, left] {$\boldsymbol{\tau}_{\mathrm{i}}(t) $}   (plant);
    % \draw[-] (plant.east) -- ++(1.65,0) node[midway, above] {$q_B(t)$};
    \draw[-] (plant.east) -- ++(4.505,0.0) node[pos=0.5, anchor=south, xshift=1.5cm, yshift=0cm]{$\ \boldsymbol{\varphi}(t),\ \dot{\boldsymbol{\varphi}}(t)\ $} |-  (LPC);
    \draw[-] (plant.east) -- ++(1.91,0.0) |-  node[pos=0.5, anchor=west, xshift=0cm, yshift=0.75cm]{$\begin{matrix} \vect{q}(t)  , \\ \boldsymbol{\omega}(t)\end{matrix}$}  (BP.north) ;

    % Input node
    \node[input, left=1cm of BP] (input) {};
    \draw[arrow, blue] (input) -- node[midway, above left] {$\vect{q}_{\mathrm{D}}$} (BP);

    \node[input, above=1.5cm of BP] (virtual_up) {};
    \draw[arrow] (virtual_up) -- node[midway, above left] {} (BP);

    \draw[arrow] (0, 1.995) -- (2.6, 1.995) node[midway, above left] {}; 
    
\end{tikzpicture}
    \caption{Controller architecture for in-flight attitude control}
    \label{MPC::fig::ControllerArchitecture}
\end{figure}

The overall control architecture is shown in Fig. \ref{MPC::fig::ControllerArchitecture}.

The Body Planner takes the latest orientation and angular velocity estimates of the body $\vect{q}(t),\boldsymbol{\omega}(t) $ and calculates the optimal body torques $\boldsymbol{\tau}_{\mathrm{B}}^{\ast}$ to stabilize the desired torso orientation $\vect{q}_{\mathrm{D}}$. The Leg Planner reads the latest joint positions and velocities $\boldsymbol{\varphi}(t),\ \dot{\boldsymbol{\varphi}}(t) $ and latest virtual torques $\boldsymbol{\tau}_{\mathrm{B}}^{\ast}$, checks whether there is a mode or phase change and calculates the optimal joint trajectories for the front right leg $\boldsymbol{\varphi}_{\mathrm{FR}}^{\ast}(\cdot),\ \boldsymbol{\tau}_{\mathrm{FR}}^{\ast}(\cdot)$, which are then mapped suitably in all the legs through the Allocation Module, as explained in Sect. \ref{MPC::sec::Allocation}. Tracking of $\boldsymbol{\varphi}_{\mathrm{i}}^{\ast}(\cdot)$ and $ \boldsymbol{\tau}_{\mathrm{i}}^{\ast}(\cdot)$ can be achieved by various controllers. In the present work, a simple position PID controller, described in (\ref{MPC::eq::PID_control}), is used in accordance with the current motor interface.

\begin{equation}
    \boldsymbol{\tau}_{\mathrm{i}} = \matr{K}_{\mathrm{P}} (\boldsymbol{\varphi}_{\mathrm{i}}^{\ast} -  \boldsymbol{\varphi}_{\mathrm{i}}(t)) 
    - \matr{K}_{\mathrm{D}} \frac{d}{dt} \left [ \boldsymbol{\varphi}_{\mathrm{i}}(t) \right ]
    -  \matr{K}_{\mathrm{I}} \int \left [ (\boldsymbol{\varphi}_{\mathrm{i}}^{\ast} -  \boldsymbol{\varphi}_{\mathrm{i}}(t) \right ]  dt 
    \label{MPC::eq::PID_control}\;,
\end{equation}
where $\matr{K}_{\mathrm{P}},\matr{K}_{\mathrm{D}},\matr{K}_{\mathrm{I}}$ is the proportional, derivative and integral gain respectively.

\section{Evaluation Studies} \label{sec::Results}
This section presents simulation and experimental results. In Sect.~\ref{sec::Implementationdetails}, implementation details particular to the conducted studies are outlined. In Sect.~\ref{sec::Simulation}, the reorientation capability and robustness of the controller are demonstrated through simulation, whereas the experimental results of Sect.~\ref{sec::Experiment} showcase the viability of the proposed controller. 

\subsection{Implementation Details} \label{sec::Implementationdetails}

The MPCs are implemented using the acados open source framework \cite{acados}, and use the HPIPM solver \cite{HPIPM}. The weights of the NMPCs (\ref{MPC::eq::BodyPlannerMPC},\ref{MPC::eq::torque_tracking}) were manually tuned.  The horizon of the Body Planner is $T_{\mathrm{H,B}}=2$~s, it has four stages with five simulation steps in each stage. The Leg Planner MPC has five stages with one simulation step per stage, while its prediction horizon is $T_{\mathrm{H,L}} = 0.1$~s.  Their update rate is $10$~Hz. The reference of the low-level controllers is updated at $50$~Hz. However, upon detecting a phase change, the leg planner immediately recalculates the trajectory, after updating its parameters. This interrupting capability of the resetting module is indicated with red in Fig.~\ref{MPC::fig::ControllerArchitecture}. For timely detection of phase transitions, the control is clocked at $1$~kHz.

The CPU used in simulations and tuning trials is an AMD Ryzen 5 5600H. On tuning tests, the Body Planner MPC is solved on average in $0.42$~ms while the Leg Planner MPC takes on average in $12.2$~ms. Due to this significant difference, only the computation time of the Leg Planner MPC is reported later. 

\subsection{Simulation Results} \label{sec::Simulation}
\begin{figure}[ht]
    \centering
    \includegraphics[width=1\linewidth]{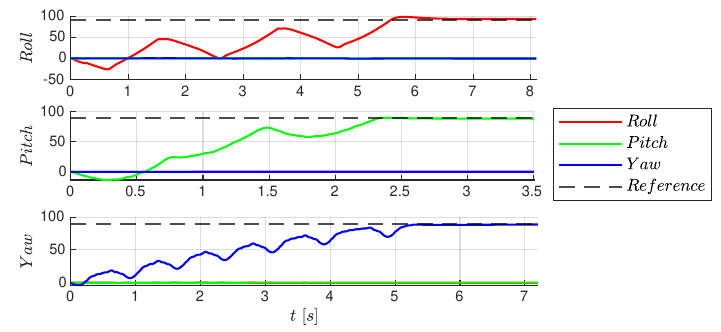}
    \caption{Single axis $90^{\circ}$ reorientation simulations}
    \label{Results::fig::SimSingleAxis}
\end{figure}

In order to tune and evaluate the proposed controller prior to its experimental validation, a simulation study took place using Drake. The quadruped is modeled as a free-floating multibody, the closure constraint (\ref{Modelling::eq::ClosureConstraint}) is implemented using a virtual spring and gravity is disabled in simulation for reasons discussed in Sect.~\ref{Modelling::sec::Dynamics}. The simulation tasks involve a $+90^{\circ}$ turn in roll, pitch, and yaw, starting from $\vect{q}_0 = [1,0,0,0]^T$, with a  convergence threshold of $5^{\circ}$. The step response plots are shown in Fig. \ref{Results::fig::SimSingleAxis}, and some quantitative results are presented in Table~\ref{Results::tab::SimSingleAxis}.

The controller successfully stabilizes all three rotations through periodic maneuvers. In all cases, off-axis deviations are minimal  (the largest being $0.8^{\circ}$), due to the  suitable mapping of the leg trajectories in the allocation module. Pitch---the most critical axis to control in forward jumps \cite{SpaceHopper}---is the fastest degree of freedom to converge, converging in $2.4$~s. Finally, the MPC of the Leg Planner is solved faster than 100 ms, the nominal update rate of the module. 

\begin{table}[ht]
\vspace{-3ex}
\centering
\caption{Simulation results of single axis $90^{\circ}$ reorientation}
\label{Results::tab::SimSingleAxis}
\begin{tabular}{r@{\hspace{1cm}}r@{\hspace{0.6cm}}r@{\hspace{.6cm}}r@{\hspace{.6cm}}}
\hline
                   & Roll & Pitch & Yaw \\ \hline
Settling Time $T_{\mathrm{ss}}$ (s)                                    & 6.3                          & 2.4                           & 5.5                     \\ 
Mean Angular Velocity $\bar{\omega}$ ($^{\circ}/$s)             & $14.9$                       & $37.5$                        & $16.1$                  \\ 
Steady State Error $e_{\mathrm{ss}}$  ($^{\circ}$)                       & $3.9$                        & $1$                           & $1.5$                   \\ 
Leg Planner MPC                                                 &                              &                               &                         \\ 
average CPU time $\mu(t_{\mathrm{CPU}})$ (ms)                   & $13$                         & $8.5$                         & $6.6$                   \\ 
standard deviation of CPU time $\sigma(t_{\mathrm{CPU}})$ (ms)     & $4.2$                        & $9$                           & $4$                     \\ 
maximum  CPU time max$(t_{\mathrm{CPU}})$ (ms)                &  $25$                        & $55$                          & $19.4$                  \\ \hline
\end{tabular}
\vspace{-2ex}
\end{table}

\begin{figure}[ht]
    \centering
    \includegraphics[width=1\linewidth]{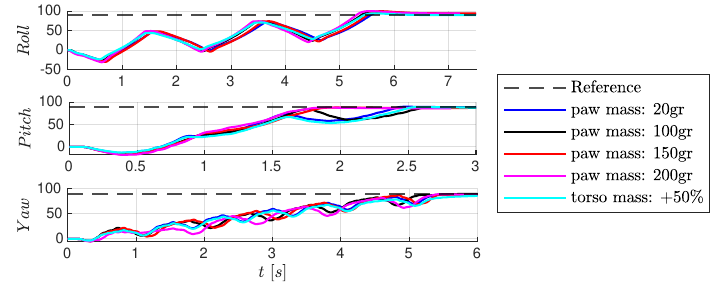}
    \caption{Simulation results of single axis $90^{\circ}$ reorientation with added mass in the paws or in the torso}
    \label{Results::fig::AblationStudy}
\end{figure}

Next, the effect of adding extra mass to the paws or the torso (e.g. due to heavier payload) is investigated. The results are shown in Fig.~\ref{Results::fig::AblationStudy}. First, it can be observed that the controller manages to stabilize the desired orientation in all cases, showcasing its robustness. Roll and yaw remain relatively unaffected. The additional mass in the paws enhances the control authority, but it also increases the system's inertia, particularly in roll, and lengthens the cycle time, as seen in yaw. However, pitch is improved with the addition of up to $150$~g in the paws. 

\subsection{Experimental Results} \label{sec::Experiment}

\begin{figure}[ht]
\centering
    \begin{subfigure}[t]{0.49\textwidth}
    \centering
    \includegraphics[width = .99\textwidth]{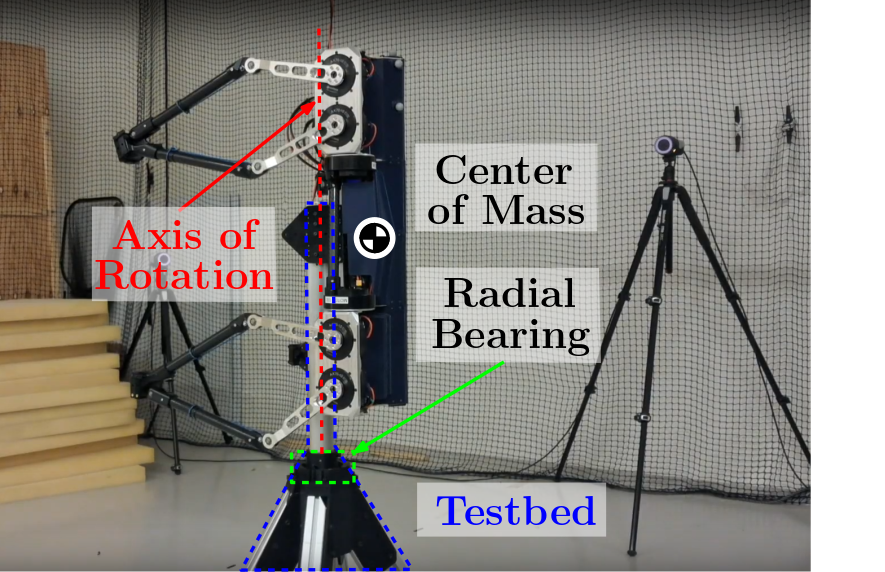}
    \caption{}
    \end{subfigure}
    \begin{subfigure}[t]{0.45\textwidth}
    \centering
    \includegraphics[width = 1\textwidth]{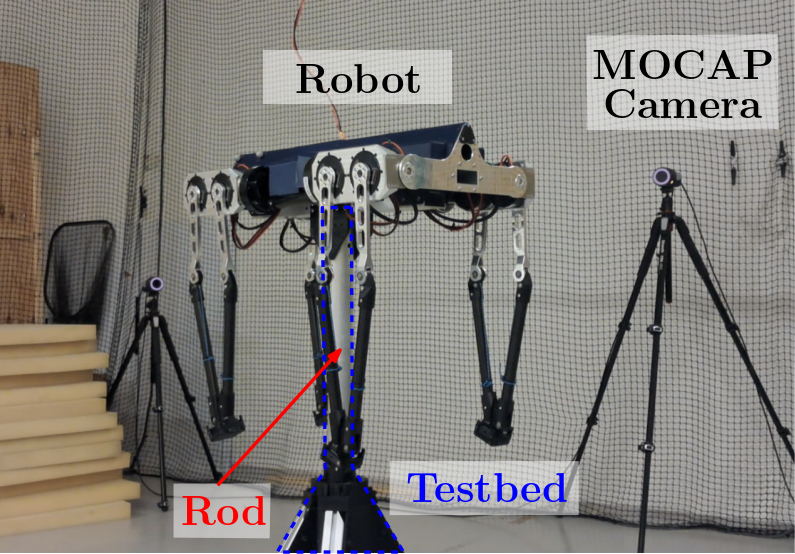}
    \caption{}
    \end{subfigure}
    \caption{\textbf{a} Experimental setup for roll. \textbf{b} Experimental setup for yaw. }
    \label{Results::fig::setup}
\end{figure}

Roll and yaw stabilization was tested on the actual robot. The experimental setup is depicted in Fig. \ref{Results::fig::setup}. The quadruped is placed on a testbed that allows rotation on a single axis, which is parallel to gravity to minimize its effect on the torso's rotation. However, the rotation is affected by bearing friction. Specifically, the bending moments caused by the non-coincidence of the center of mass and the axis of rotation compromise the low-friction characteristics of the bearings.

Olympus is equipped with CubeMars AK 80-9 and AK70-10 brushless DC motors (BLDC) and employs an Nvidia OrinX, with an Arm Cortex-A78AE CPU, for onboard computations. Important design parameters\footnote{The reported inertia values are calculated in the configuration shown in Fig.~\ref{Modelling::fig::collision_geometry}.} are presented in Table \ref{Results::tab::OlympusDetails}, while further design details can be found in \cite{olsen2023martian}. Orientation feedback is provided by a motion capture system. Using ROS, the controller, state estimation, and motor control nodes were deployed on the onboard computer, while the robot was operated from a separate computer. Communication between the systems was established via WiFi.

\begin{figure}[h]
    \vspace{-1.5ex}
    \centering
    \includegraphics[width=0.9\linewidth]{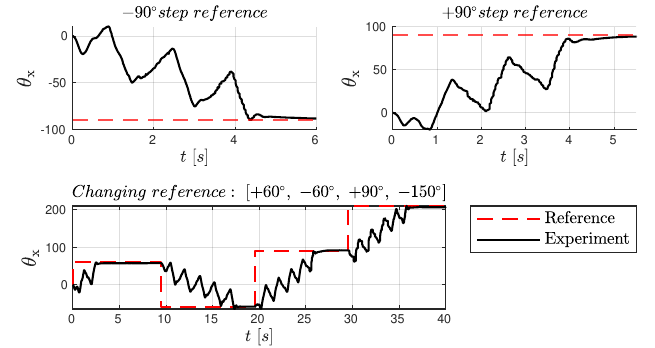}
    \caption{Roll Experimental Results}
    \label{Results::fig::RollExperiment}
\end{figure}

\begin{figure} [ht]
    \vspace{-1.5ex}
    \centering
    \includegraphics[width=.9\linewidth]{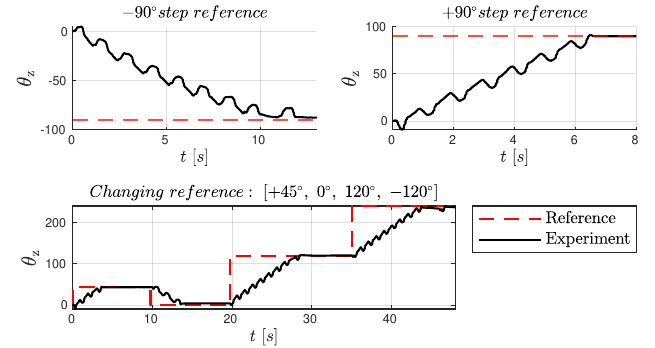}
    \caption{Yaw Experimental Results}
    \label{Results::fig::YawExperiment}
\end{figure}

Two kinds of experiments took place. A response to a step input reference of $\pm90^{\circ}$ and a response to a changing reference. The results of the experiments are shown in Figs. \ref{Results::fig::RollExperiment} and \ref{Results::fig::YawExperiment}, and some indicative metrics are included in Table \ref{Results::tab::ExperimentalResults}.

\begin{table}[H]
    \centering
    \vspace{-3ex}
    \begin{minipage}[t]{0.32\textwidth}
        \caption{Olympus design parameters}
        \label{Results::tab::OlympusDetails}
        \begin{tabular}{r@{\hspace{0.75cm}}r}
                                                                \hline
            Quantity                        & Value         \\  \hline
            Length  (mm)                    & $679$         \\ 
            Width  (mm)                     & $400$         \\ 
            Height (mm)                     & $570$         \\ 
            $\matr{I}_{\mathrm{x}}$ ($\mathrm{kg \ m^2}$)       & $0.49$        \\ 
            $\matr{I}_{\mathrm{y}}$ ($\mathrm{kg \ m^2}$)       & $0.92$        \\ 
            $\matr{I}_{\mathrm{z}}$ ($\mathrm{kg \ m^2}$)       & $1.01$        \\ 
            Body mass (kg)                  & $5.68$        \\ 
            Leg mass (kg)                   & $2.02$        \\  \hline
        \end{tabular}
    \end{minipage}
    \hfill % add some horizontal space between the tables
    \begin{minipage}[t]{0.6\textwidth}
        \caption{Experimental results of single axis $90^{\circ}$ reorientation}
        \label{Results::tab::ExperimentalResults}
            \begin{tabular}{r@{\hspace{0.5cm}}r@{\hspace{0.2cm}}r@{\hspace{.5cm}}r@{\hspace{.2cm}}r@{\hspace{.3cm}}}
                                                                                                                                        \hline
                Metric                                & \multicolumn{2}{c}{Roll}                & \multicolumn{2}{c}{Yaw}               \\ 
                                                      & $+90^{\circ}$       & $-90^{\circ}$     & $+90^{\circ}$     & $-90^{\circ}$     \\ \hline
                $T_{\mathrm{ss}}$ (s)                        & 5                 & 5.2               & 6.5                 & 12                \\ 
                $\bar{\omega}$ ($^{\circ}/$s)         & $17.5$            & $16.8$            & $13.9$              & $7.3 $            \\ 
                $e_{\mathrm{ss}}$  ($^{\circ}$)                & $2.6$             & $2.6$             & $0.8$               & $2.2$             \\ 
                Leg Planner MPC                       &                   &                   &                     &                   \\
                $\mu (t_{\mathrm{CPU}})$  (ms)                   & $14$              & $12$              & $19$                & $14$              \\ 
                $\sigma (t_{\mathrm{CPU}})$ (ms)                    & $5.4$             & $4.4$             & $8.6$               & $6.8$             \\ 
                $max (t_{\mathrm{CPU}} )$ (ms)                 & $35$              & $25$              & $63$                & $43$              \\ \hline
            \end{tabular}
    \end{minipage}
    % \vspace{-3ex}
\end{table}

The controller manages to stabilize the system despite the external disturbances (gravity and rotational friction). Roll stabilizes in around $5$~s for $\pm 90^{\circ}$ rotations, while yaw requires $6.5$~s and $12$~s for the $+90^{\circ}$ and $-90^{\circ}$ rotations respectively. The difference in yaw can be attributed to imbalance of the rod and the manual tuning of the controller, which has created some anisotropy. In both cases, when tracking changing references, the controller selected the shortest turning path in the final maneuver by continuing to turn in the positive direction. In Figs. \ref{Results::fig::RollExperiment} and \ref{Results::fig::YawExperiment}, an equivalent orientation is shown as the last reference to demonstrate the convergence.

\FloatBarrier

\section{Conclusions} \label{sec::Conclusion}
In this work, the modeling and attitude control of a jumping quadruped were investigated. The proposed controller decouples torso orientation planning from leg motion generation, successfully stabilizing the system's attitude and avoiding self-collisions. Simulations and experiments showcase the viability of the method. 

However, a set of limitations remains to be solved. First, the large number of tuning parameters involved makes achieving satisfactory stabilizing performance a considerable challenge. Second, the complexity of the leg model severely limits the tractable length of the prediction horizon of the Leg Planner and, in turn, its optimality. Finally, the proposed controller requires improvements when it comes to stabilizing composite orientations due to the employed discretization of turning maneuvers in roll, pitch, and yaw modes as well as the tuning of the MPC parameters to single axis rotations. 

To address these limitations, supervised learning can be used to generalize the control system, allowing for dynamic adjustment of MPC parameters to accommodate composite 3D rotations. Alternatively, hyperparameter optimization tools can simplify the tuning process. 
Overall, this work addresses key aspects of model-based attitude control for jumping legged systems and offers a framework for future advancements.

\subsubsection*{Acknowledgements.}
This work was partially supported by the Research Council of Norway under Grant Agreement 338694.

\FloatBarrier
\bibliographystyle{ieeetr}
\bibliography{refrences}

\begin{thebibliography}{10}

\bibitem{SpaceHopper}
A.~Spiridonov, F.~Buehler, M.~Berclaz, V.~Schelbert, J.~Geurts, E.~Krasnova, E.~Steinke, J.~Toma, J.~Wuethrich, R.~Polat, W.~Zimmermann, P.~Arm, N.~Rudin, H.~Kolvenbach, and M.~Hutter, ``{SpaceHopper}: A small-scale legged robot for exploring low-gravity celestial bodies,'' {\em arXiv preprint arXiv:2403.02831}, 2024.

\bibitem{vaquero2024eels}
T.~S. Vaquero, G.~Daddi, R.~Thakker, M.~Paton, A.~Jasour, M.~P. Strub, R.~M. Swan, R.~Royce, M.~Gildner, P.~Tosi, {\em et~al.}, ``Eels: Autonomous snake-like robot with task and motion planning capabilities for ice world exploration,'' {\em Science Robotics}, vol.~9, no.~88, p.~eadh8332, 2024.

\bibitem{olsen2023martian}
J.~A. Olsen and K.~Alexis, ``Martian lava tube exploration using jumping legged robots: A concept study,'' {\em arXiv preprint arXiv:2310.14876}, 2023.

\bibitem{LavaTubesReport}
A.~Daga, A.~Carlton, J.~Burke, I.~Crawford, R.~Leveille, S.~Simon, and L.~T. Tan, ``Lunar and martian lava tube exploration as part of an overall scientific survey,'' 2013.

\bibitem{Spacebok1}
H.~Kolvenbach, E.~Hampp, P.~Barton, R.~Zenkl, and M.~Hutter, ``Towards jumping locomotion for quadruped robots on the moon,'' in {\em 2019 IEEE/RSJ International Conference on Intelligent Robots and Systems (IROS)}, pp.~5459--5466, 2019.

\bibitem{Spacebok2}
N.~Rudin, H.~Kolvenbach, V.~Tsounis, and M.~Hutter, ``Cat-like jumping and landing of legged robots in low gravity using deep reinforcement learning,'' {\em IEEE Transactions on Robotics}, vol.~38, no.~1, pp.~317--328, 2022.

\bibitem{OlsenDesign}
J.~A. Olsen and K.~Alexis, ``Design and experimental verification of a jumping legged robot for martian lava tube exploration,'' {\em 2023 21st International Conference on Advanced Robotics (ICAR)}, pp.~452--459, 2023.

\bibitem{westre2024model}
A.~Westre, J.~A. Olsen, and K.~Alexis, ``Model predictive attitude control of a jumping-and-flying quadruped for planetary exploration,'' in {\em 2024 IEEE Aerospace Conference}, pp.~1--12, IEEE, 2024.

\bibitem{cheetah_falling}
V.~Kurtz, H.~Li, P.~M. Wensing, and H.~Lin, ``Mini cheetah, the falling cat: A case study in machine learning and trajectory optimization for robot acrobatics,'' in {\em 2022 International Conference on Robotics and Automation (ICRA)}, pp.~4635--4641, 2022.

\bibitem{Nasa_ADCS}
S.~R. Starin and J.~S. Eterno, ``Attitude determination and control systems,'' tech. rep., Nasa Goddard Space Flight Center, 2011.

\bibitem{virtual_manipulator}
Z.~Vafa and S.~Dubowsky, ``On the dynamics of manipulators in space using the virtual manipulator approach,'' in {\em Proceedings. 1987 IEEE International Conference on Robotics and Automation}, vol.~4, pp.~579--585, 1987.

\bibitem{manipulator_spacecraft}
I.~Tortopidis and E.~Papadopoulos, ``On point-to-point motion planning for underactuated space manipulator systems,'' {\em Robotics and Autonomous Systems}, vol.~55, no.~2, pp.~122--131, 2007.

\bibitem{falling_cat}
J.~Cao, S.~Wang, X.~Zhu, and L.~Song, ``A review of research on falling cat phenomenon and development of bio-falling cat robot,'' in {\em 2022 WRC Symposium on Advanced Robotics and Automation (WRC SARA)}, pp.~160--165, 2022.

\bibitem{quaternion_kf}
J.~Sol{\`a}, ``Quaternion kinematics for the error-state kalman filter,'' {\em arXiv preprint arXiv:1711.02508}, 2017.

\bibitem{Goldstein2001-iq}
H.~Goldstein, C.~P. Poole, and J.~L. Safko, {\em Classical Mechanics}.
\newblock Upper Saddle River, NJ: Pearson, 3~ed., June 2001.

\bibitem{5-bar_kinematics}
G.~Campion, Q.~Wang, and V.~Hayward, ``The pantograph mk-ii: a haptic instrument,'' in {\em 2005 IEEE/RSJ International Conference on Intelligent Robots and Systems}, pp.~193--198, 2005.

\bibitem{underactuated}
R.~Tedrake, {\em Underactuated Robotics}.
\newblock 2023.

\bibitem{FeatherstoneDynamics}
R.~Featherstone, {\em Rigid body dynamics algorithms}.
\newblock New York, NY: Springer, 2~ed., Nov. 2007.

\bibitem{drake}
R.~Tedrake and the Drake Development~Team, ``Drake: Model-based design and verification for robotics,'' 2019.

\bibitem{acados}
R.~Verschueren, G.~Frison, D.~Kouzoupis, J.~Frey, N.~van Duijkeren, A.~Zanelli, B.~Novoselnik, T.~Albin, R.~Quirynen, and M.~Diehl, ``acados -- a modular open-source framework for fast embedded optimal control,'' {\em Mathematical Programming Computation}, Oct 2021.

\bibitem{HPIPM}
G.~Frison and M.~Diehl, ``{HPIPM}: a high-performance quadratic programming framework for model predictive control,'' 2020.

\end{thebibliography}

\end{document}